\documentclass[jounal]{acmsiggraph}

\usepackage{booktabs} 

\usepackage[ruled]{algorithm2e} 
\usepackage{amsmath}
\usepackage{amssymb}
\usepackage{mathrsfs}
\usepackage[british,UKenglish,USenglish,english,american]{babel}
\usepackage{booktabs}
\usepackage{color}
\usepackage{subfigure}
\usepackage{array}

\usepackage{epsfig}
\usepackage{graphicx}
\usepackage{ifthen}

\usepackage{hyperref}
\usepackage{array}
\usepackage{color}
\usepackage{amsmath}
\usepackage{float}

\newcommand{\commentout}[1]{} 
{}

\newcolumntype{C}[1]{>{\centering\let\newline\\\arraybackslash\hspace{0pt}}m{#1}}
\newcommand{\RNum}[1]{\uppercase\expandafter{\romannumeral #1\relax}}

\DeclareMathOperator*{\argmax}{arg\,max}

\usepackage{array}
\newcolumntype{L}[1]{>{\raggedright\let\newline\\\arraybackslash\hspace{0pt}}m{#1}}
\newcolumntype{C}[1]{>{\centering\let\newline\\\arraybackslash\hspace{0pt}}m{#1}}
\newcolumntype{R}[1]{>{\raggedleft\let\newline\\\arraybackslash\hspace{0pt}}m{#1}}

\title{MichiGAN: Multi-Input-Conditioned Hair Image Generation for Portrait Editing}

\author{Zhentao Tan$^1$, Menglei Chai$^2$, Dongdong Chen$^3$, Jing Liao$^4$, Qi Chu$^1$, Lu Yuan$^3$, Sergey Tulyakov$^2$ and Nenghai Yu$^1$
	\\
	$^1$University of Science and Technology of China,
	$^2$Snap Inc.,
	$^3$Microsoft Cloud AI,
	$^4$City University of Hong Kong
}

\pdfauthor{Tan, Chai, Chen, Liao, Chu, Yuan, Tulyakov and Yu}

\keywords{interactive portrait editing; conditional hair image generation; generative adversarial networks}

\begin{document}

\teaser{
  \centering
  \footnotesize
\includegraphics[width=1.0\linewidth]{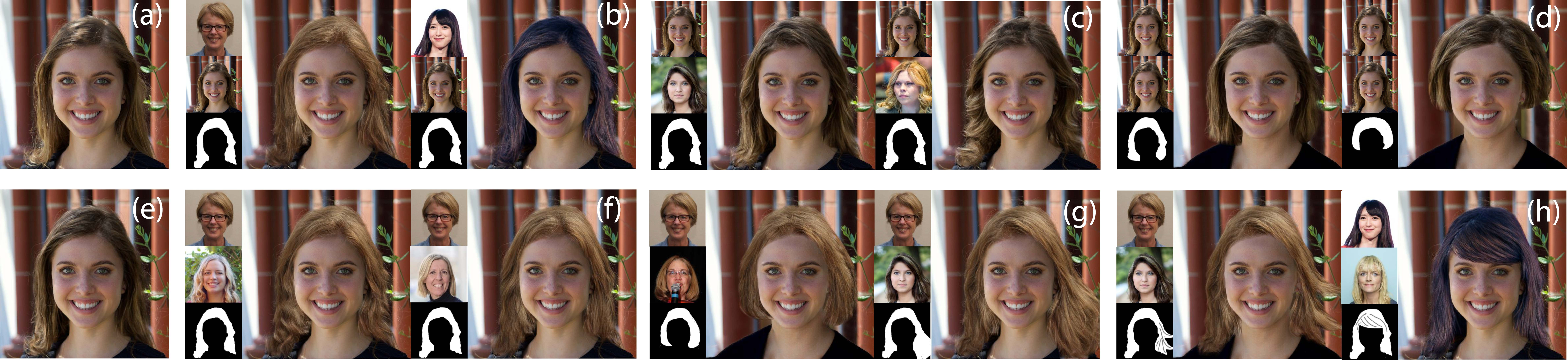}\\
\caption{Given an input portrait image (a), MichiGAN is capable of enabling multiple input conditions for disentangled hair manipulation. It not only supports editing one attribute individually including appearance (b), structure (c), and shape (d) while keeping the background unchanged, but also allows manipulating multiple attributes (f\&g) jointly. The users can also leverage our interactive system to create more customized hair manipulation results (h). (e) is the reconstruction result when all the condition inputs are from (a) itself. Original images courtesy of stephen davis, Eric E Castro, LeafBlue, Pawel Loj, O'Reilly Internal, Lydia Liu, Faculty of Social and Educational Sciences, NWABR, and Zero Emission Resource Organisation.}
\label{fig:teaser}
}

\maketitle

\begin{abstract}
	Despite the recent success of face image generation with GANs, conditional hair editing remains challenging due to the under-explored complexity of its geometry and appearance. In this paper, we present MichiGAN (Multi-Input-Conditioned Hair Image GAN), a novel conditional image generation method for interactive portrait hair manipulation. To provide user control over every major hair visual factor, we explicitly disentangle hair into four orthogonal attributes, including shape, structure, appearance, and background. For each of them, we design a corresponding condition module to represent, process, and convert user inputs, and modulate the image generation pipeline in ways that respect the natures of different visual attributes. All these condition modules are integrated with the backbone generator to form the final end-to-end network, which allows fully-conditioned hair generation from multiple user inputs. Upon it, we also build an interactive portrait hair editing system that enables straightforward manipulation of hair by projecting intuitive and high-level user inputs such as painted masks, guiding strokes, or reference photos to well-defined condition representations. Through extensive experiments and evaluations, we demonstrate the superiority of our method regarding both result quality and user controllability. The code is available at \url{https://github.com/tzt101/MichiGAN}.
\end{abstract}

\section{Introduction}
Human hair is so delicate, variable, and expressive that it constantly plays a unique role in depicting the subject in a face image. Given its diversity and flexibility, the urge to manipulate hair in a portrait photo, with a full spectrum of user control, has been long coveted to facilitate truly personalized portrait editing -- from extending a particular hair wisp, altering the structure of a local region, to entirely transferring the hair color and appearance.

However, unlike most other parts of the human face, its peculiar geometry and material make hair extremely difficult to analyze, represent, and generate. The complexity of visual perception comes from multiple factors that the user may intend to edit or preserve. If we take a closer look, by roughly decomposing the hair region into its corresponding mask shape and color appearance, a few key aspects can be identified that are crucial to the visual quality. Shape-wise, the intricate mask boundaries, and varying opacity can lead to challenges of precise hair shape control and seamless background blending. Appearance-wise, the anisotropic nature of hair fibers tends to entangle both the strand flows and the material properties, in which the former renders the major local structures while the latter determines the color palette and inter-strand variations in a globally consistent manner. Therefore, to achieve controllable hair manipulation, we need to have the system not only capable of generating photo-realistic hair images, but also able to bridge various user controls and use them to condition the hair generation in ways that respect the natures of different visual factors.

Fortunately, the latest advances in deep neural networks, or more specifically, generative adversarial networks (GAN), have driven significant progress on human face synthesis. Recent successes, such as ProgressiveGAN~\cite{karras2018progressive} and StyleGAN~\cite{karras2019style}, already make it possible to produce highly realistic human face, including hair, from a random latent code. Yet conditioned generation, the heart of interactive face editing, is thought to be more challenging and less explored. Despite the recent efforts on conditioned face generation with semantic maps~\cite{lee2019maskgan} or sketches~\cite{jo2019sc}, we are still far from reaching high-quality hair editing that is both fully-controllable and user-friendly.

In this work, we present \textit{MichiGAN} (Multi-Input-Conditioned Hair Image GAN), a novel conditional hair image generation method for interactive portrait manipulation. To achieve so, we explicitly disentangle the information of hair into a quartet of attributes -- \textit{shape}, \textit{structure}, \textit{appearance}, and \textit{background}, and design deliberate representations and conditioning mechanisms to enable orthogonal control within the image generation pipeline.

To this end, we propose three distinct condition modules for these attributes according to their particular perceptual characteristics and scales. Specifically, spatially-variant attributes of \textit{shape} and \textit{structure} are represented as fuzzy semantic masks and weighted orientation maps, respectively, which modulate the network with the conditional normalization structures~\cite{huang2017arbitrary,park2019semantic}.
More globally, \textit{appearance} is encoded through our mask-transformed feature extracting network, which acts as the latent code input to the very beginning of the network and bootstraps the generator with appearance styles guided by arbitrary references. In addition, a \textit{background} encoder is placed parallel to the generation branch, which keeps background intact by progressively injecting features into the generator in a mask-aware way without loss of generation capability.

By integrating all these condition modules, we have an end-to-end image generation network that can provide complete control over every major hair attribute. Tailored for such a complicated network, we further develop a data preparation and augmentation pipeline for disentangling hair into compatible training data pairs, together with a combination of novel loss functions to achieve smooth and robust training. By extensively evaluating MichiGAN on a wide range of in-the-wild portrait images and comparing it with state-of-the-art conditioned image generation methods and alternative network designs, we demonstrate the superiority of the proposed method on both result quality and controllability.

Moreover, we build an interactive portrait hair editing system based on MichiGAN. It enables straightforward manipulation of hair by projecting intuitive and high-level user inputs such as painted masks, guiding strokes, or reference photos to well-defined condition representations. By using this hair editing interface, users are allowed to modify the hair shape through mask painter, alter or inpaint local hair structure with sparse stroke annotations, replace hair appearance using another hair reference, or even edit multiple attributes all at once concurrently.

In summary, the main contributions of this work are:
\begin{itemize}
\item An \textit{explicit disentanglement of hair visual attributes}, and a set of \textit{condition modules} that implement the effective condition mechanism for each attribute with respect to its particular visual characteristics;
\item An \textit{end-to-end conditional hair generation network} that provides complete and orthogonal control over all attributes individually or jointly;
\item An \textit{interactive hair editing system} that enables straightforward and flexible hair manipulation through intuitive user inputs.
\end{itemize}

The training code, pre-trained models, and the interactive system are made publicly available to facilitate future research (\href{https://github.com/tzt101/MichiGAN.git}{GitHub}).

\section{Related Work}

\paragraph{Portrait Manipulation}
Portrait image manipulation is a long-standing while extensively studied area in both vision and graphics communities. Early approaches tackle specific application-driven problems with conventional image editing techniques, such as face swapping~\cite{bitouk2008face}, face shape beautification~\cite{leyvand2008data}, and portrait enhancement~\cite{joshi2010personal}. More recently, the popularization of image-based 3D face reconstruction~\cite{blanz1999morphable} enables lots of 3D-aware portrait manipulation methods, on expression editing~\cite{yang2011expression}, relighting~\cite{wang2009face}, aging~\cite{shlizerman2014illumination}, camera manipulation~\cite{fried2016perspective}, geometry reshaping~\cite{cao2014facewarehouse}, identity replacement~\cite{dale2011video,shlizerman2016transfiguring}, and performance reenactment~\cite{thies2016face2face}. However, the limited fitting accuracy and model quality of face proxy tends to affect the result realism. Thanks to the recent progress on deep neural networks, high-fidelity portrait manipulation has been made feasible in multiple scenarios, including sketch-based generation~\cite{lee2019maskgan,jo2019sc}, attribute editing~\cite{shu2017neural,geng2019guided}, view synthesis~\cite{nagano2019deep}, and relighting~\cite{sun2019single}.

\paragraph{Hair Modeling and Editing}
As a critical yet challenging component of the human face, hair is also of great interest to researchers. Most hair acquisition pipelines focus on capturing hair geometry from various image inputs, including multi-view~\cite{paris2008hair,luo2013structure,hu2014robust,zhang2017data} and single-view~\cite{chai2013dynamic,hu2015single,chai2016autohair,zhou2018hairnet,saito2018hair,liang2018video}. Utilizing the coarse hair geometry approximated from single-view images, a few hair editing methods are proposed on hair transferring~\cite{chai2012single}, animation~\cite{chai2013dynamic}, relighting~\cite{chai2015high}, and morphing~\cite{weng2013hair}. But the lack of in-depth understanding and manipulation approaches to many of key visual factors greatly compromise the result quality and editing flexibility. The highly realistic portrait synthesis results from recent GAN-based image generated methods~\cite{karras2019style} have shown great potentials in bridging this quality gap. However, conditioned image generation is generally thought to be much more difficult than generation from random codes. Although some recent work~\cite{wei2018real,lee2019maskgan,jo2019sc,qiu2019two} have achieved progress on hair generation conditioned by limited types of inputs, these methods are unfortunately not intuitively controllable and universally applicable. In this paper, we propose a set of disentangled hair attributes to cover the full spectrum of visual hair factors, and for each attribute, we design a conditional module that could effectively control the generator with user-friendly inputs. This completely-conditioned and controllable hair generation has not been achieved before.

\paragraph{GAN-based Image Generation}
Thanks to the powerful modeling capacity of deep neural networks, varieties of generative models have been proposed to learn to synthesize images in an unconditional setting. Typical methods include generative adversarial networks (GANs) \cite{goodfellow2014generative,metz2016unrolled,mao2017least} and variation auto-encoders (VAE) \cite{doersch2016tutorial,sonderby2016ladder}. Compared to VAE, GANs are demonstrated to be more popular and capable of modeling fine-grained details. Starting from a random latent code, GANs can generate fake images with the same distribution as natural images in a target domain. The famous work ProgressiveGAN ~\cite{karras2018progressive} first leveraged a progressive generative network structure that can generate very highly realistic facial images, including hair. By further incorporating the latent and noise information into both the shallow and deep layers, StyleGAN \cite{karras2019style} further improved the generation quality significantly. Our work is built on GANs but aims for the conditional image generation task, which is much more challenging due to multiple elaborate conditions for controlling desired hair synthesis.
Utilizing these pre-trained GAN networks, many recent works \cite{abdal2019image,bau2019semantic} achieve convincing image editing results (including portrait) by manipulating the embedded features in the latent spaces of these networks.

\paragraph{Conditional Image Generation}
Many recent works leverage conditional GAN (cGAN) for image generation, conditioned on different types of inputs, such as category labels~\cite{mirza2014conditional,odena2017conditional,brock2019large}, text~\cite{reed2016generative,zhang2017stackgan,yin2019semantics}, or images~\cite{isola2017image,wang2018high,park2019semantic,jo2019sc,lee2019maskgan}. Among these various forms, image-conditioned generation, i.e., image-to-image translation, lies in the heart of interactive image editing. Pix2Pix~\cite{isola2017image} proposes a unified framework for different applications, such as transforming a semantic segmentation mask or sketch to a photo-realistic image. It is further improved by Pix2PixHD~\cite{wang2018high} with multi-scale generators for high-resolution image generation. SPADE~\cite{park2019semantic} proposes spatially-adaptive normalization for better converting segmentation mask into photo-realistic image.
A few works~\cite{singh2019finegan,yao20183d,hong2018learning} propose disentangled image generation, which share similar motivations as ours.
Recently, some works have made efforts on interactive facial image editing using conditional GAN. For example, MaskGAN~\cite{lee2019maskgan} focuses on facial image generation based on fine-grained semantic masks. SC-FEGAN~\cite{jo2019sc} extends the conditional inputs to sketches and color, which allows users to control the shape and color of the face. Despite the fact that great progress has been made in conditional image generation, we are still far from reaching high-quality hair editing that is both fully-controllable due to its complexity. In this paper, we focus on hair image generation conditioned on multiple attributes that can enable user-friendly interactive hair editing.

\begin{figure*}[ht]
  \centering
  \includegraphics[width=1.58\columnwidth]{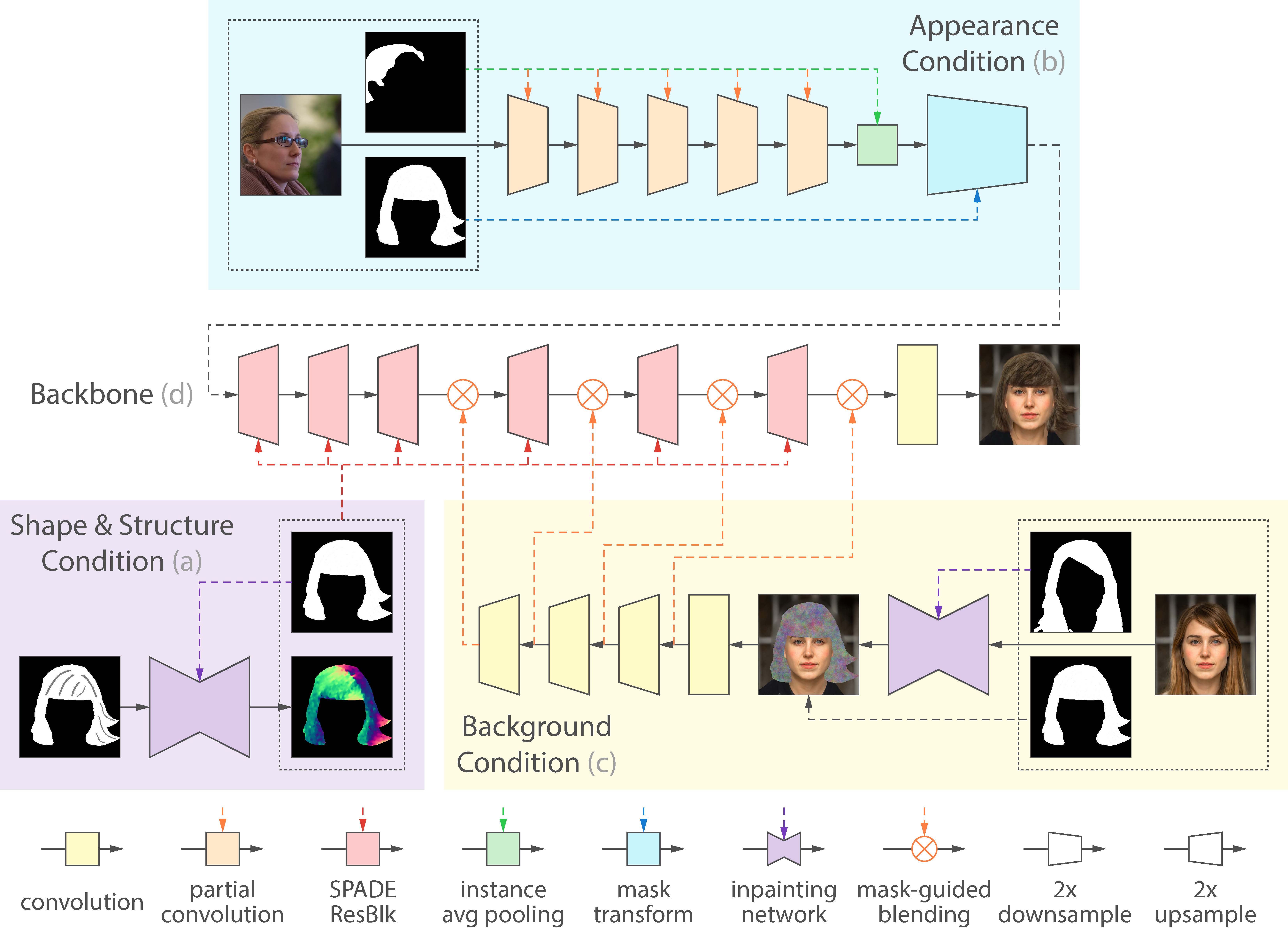}
  \caption{\textbf{The overall pipeline of MichiGAN.} It consists of a backbone generation network (d) and three condition modules for shape \& structure (a), appearance (b), and background (c). Original images courtesy of Thomas Hawk and Great Place to Work Deutschland.}
  \label{fig:pipelines}
\end{figure*}

\section{Overview}
\label{sec:overview}

Given an input portrait image $I$, our MichiGAN aims at performing conditional hair editing to the hair region according to various user control inputs, while keeping the background unchanged. The output is denoted as $I_{out}$.

As illustrated in Fig.~\ref{fig:pipelines}, the proposed pipeline consists of a backbone generation network and three condition modules, which modulate the image generation network in various ways that will be introduced later. Thanks to these three condition modules, the network provides orthogonal controls over four different hair attributes: shape, structure, appearance, and background; represented as semantic hair mask $M$, dense orientation map $O$, hair appearance reference $I_{ref}$, and the background region of the input image $I$, respectively. By combining all condition inputs, we can formulate the generation process as a deterministic function $\mathcal{G}$ as:
\begin{equation}
    I_{out} = \mathcal{G} (M, O, I_{ref}, I).
\end{equation}

These conditions, except the background, can either come from new sources in the interactive system or simply be extracted from the target/reference images. Therefore, users are not just allowed only to edit one attribute but fix all others, but can change multiple of them altogether at the same time.

We will elaborate on the technical method and the entire system in the next three sections, which are organized as follows. We first introduce the representations of our conditions, structures, and mechanisms of their corresponding condition modules (Sec.~\ref{sec:modules}). Then, we present the backbone generation network of MichiGAN that integrates all condition modules, and discuss its training strategies (Sec.~\ref{sec:generator}). Finally, our interactive hair editing system, built upon MichiGAN, is introduced together with various interaction methods that enable user-friendly control over all attributes (Sec.~\ref{sec:system}).

\section{Condition Modules}
\label{sec:modules}

In this section, we introduce our condition modules for controlling specific semantic attributes of hair during editing. In light of different natures of target attributes, their corresponding condition modules should vary as well. Here we propose three distinct modules to handle four types of main attributes, and also introduce the condition representation, component network design, and integration to the backbone network for each of them.

\subsection{Shape \& Structure}
\label{sec:modules_1}

\paragraph{Hair shape.} 
We represent the hair shape as the 2D binary mask of its occupied image region, similar to those image generation methods conditioned by semantic label maps~\cite{park2019semantic}.

However, unlike other types of objects, the visual translucency and complexity of hair boundaries make binary mask inevitably a coarse approximation of the shape. To robustly generate realistic shape details and natural blending effects from rough masks either estimated with semantic segmentation networks or interactively painted by users, we relax the strictness of the condition by adopting fuzzy mask conditioning, to allow a certain level of flexibility within the boundary area. To achieve so, we dilate or erode the mask by a random width for each data pair during training, to decouple the boundary details with the accurate shape contours to some extent.

\paragraph{Hair structure.} 
Regarding the structure of the target object, despite its generality, a mask shape is usually too sparse and ambiguous to condition the generation, which is particularly true for hair. But luckily, hair is such a special object with strong anisotropy and homogeneity that you can always depict the internal structure of hair with the strand flows uniformly distributed within the region. Therefore, we use a dense orientation map to represent the hair structure, which defines the 2D orientation angle for each pixel.

We use the oriented filter to estimate confident pixel-wise orientation maps, which has been widely used in hair reconstruction~\cite{paris2008hair} and rendering~\cite{wei2018real}. Given a bank of oriented filter kernels $\{K_{\theta}\}$ (we use $32$ Gabor filters in our implementation), with $\theta \in [0, \pi)$, we estimate the orientation label map $o'$ of image $I$, and its associated confidence $w$, at pixel position $p$ as:
\begin{equation}
    o'_p = \argmax_{\theta} \left |(K_{\theta} \otimes I)_p \right |,
\label{eq:F_filter_theta}
\end{equation}
\begin{equation}
    w_p = \max_{\theta} \left |(K_{\theta} \otimes I)_p \right |,
\label{eq:F_filter_conf}
\end{equation}
where $\otimes$ represents the convolution operator.

Continuous rotation representation is non-trivial for neural networks~\cite{zhou2019on}. The orientation label map $o'$ contains discrete angle values in  $[0, \pi)$, which are actually non-continuous since orientation is cyclic, i.e., orientation $0$ equals to $\pi$. Thus, direct use of the orientation label map will introduce ambiguity issues. We convert the label map $o'$ to two-channel continuous orientation map $O'$ for each pixel $p$ as follows:
\begin{equation}
\label{eq:F_filter_theta2}
    O'_p = [\cos(2 \cdot o'_p), \sin(2 \cdot o'_p)],
\end{equation}
and do the backward conversion as:
\begin{equation}
\label{eq:F_filter_theta3}
    o'_p = \arctan(O'_p) / 2.
\end{equation}

We use the orientation map in two ways during both network training and inference to achieve structure control:
\begin{itemize}
    \item \textbf{Condition input.} We use a dense orientation map as the input to the shape \& structure condition module. However, the raw output $O'$ from orientated filters may suffer from inaccurate results due to the existence of noise and dark regions. Therefore, we calculate the final dense orientation map $O$ by applying one pass of Gaussian smoothing to $O'$, weighted locally by confidence $w$.
    \item \textbf{Training supervision.} To explicitly enforce supervision on the structure of the result, we also propose a novel structural loss during training. Basically, we formulate the orientation estimation steps as one differentiable layer in the network following Eq.~(\ref{eq:F_filter_theta},\ref{eq:F_filter_conf}) after output $I_{edit}$, and measure per-pixel L1 orientation distance in the hair region weighted by confidence $w$. The detailed formulation is discussed in Sec.~\ref{sec:generator_2}.
\end{itemize}

We generate condition input of both shape and structure by stacking $M$ and $O$ together. Regarding the spatial-variant natures of them, we adopt spatially-adaptive normalization~\cite{park2019semantic} to achieve conditioning, as shown at Fig.~\ref{fig:pipelines} (a).

\subsection{Appearance}
\label{sec:modules_2}

Different from shape or structure, hair appearance describes the globally-consistent color style that is invariant to specific shape or structure. It includes multiple factors such as the intrinsic albedo color, environment-related shading variations, and even the granularity of wisp styles. Due to its global consistency, appearance can be compactly represented. Given the complexity of natural hair styles, instead of generating the appearance features from scratch, we address hair appearance as a style transfer problem, which extracts style from the reference image and transfers it to the target.

To force disentanglement of hair appearance, we design the appearance condition module such that it operates globally and is not susceptible to local shape and structure. Furthermore, it should be mask-invariant to be able to generalize to any hair mask at test time. 

The detailed architecture of our appearance condition network is shown in Fig.~\ref{fig:pipelines} (b). 
To this end, we have the following key ideas that are critical to the result quality:
\begin{itemize}
    \item To ensure that the encoded appearance features focus only on the hair region and are not influenced by background information, we adopt partial convolution~\cite{liu2018image} with the hair mask $M_{ref}$ of the reference image;
    \item In order to capture the global hair appearance and discard spatial-variant information as much as possible, we add an instance-wise average pooling layer to the output of the last partial convolution, within the reference hair region $M_{ref}$;
    \item To align the appearance code with the target hair shape, and uniformly distribute the extracted appearance feature, we spatially duplicate the reference feature to every pixel within the target shape $M$ to produce the appearance map $A$.
\end{itemize}

Therefore, the final appearance map $A$ is calculated as:
\begin{equation}
    A = \frac{\sum (A'_{ref} * M_{ref})} {\sum M_{ref}} \odot M,
\label{eq:F_refp}
\end{equation}
where $A'_{ref}$ is the output of the last partial convolution, $*$ denotes element-wise multiplication, and $\odot$ denotes the duplication operator.

The output of our mask-transformed feature extracting network, the appearance map $A$, is used as the latent code input to the very beginning of the backbone generation network to bootstrap the generator with specific appearance styles instead of random noises.

\subsection{Background}
\label{sec:modules_3}

Finally, we need to keep all non-hair pixels (namely the background) untouched, so that we can restrict the modification to the target hair region $M$ only. This is not as trivial as it appears~\cite{bau2019semantic}, since a) it wastes lots of network capability and greatly limits the foreground generation flexibility, if we force the GAN to reconstruct the same background content; and b) it introduces severe blending artifacts, if we only generate the foreground region and then blend it with the original background as a post-processing step.

To pass necessary background information to the generation network in a gentle way that it can easily reconstruct the background pixels and won't be distracted by the additional constraints, we design our background condition module to be an encoder network placed parallel to the last few layers in the backbone network, as shown at Fig.~\ref{fig:pipelines} (c). The encoder tries to produce multi-level feature maps that could be easily reused by the generator to reconstruct the background. The encoded feature after each convolution layer is designed to have the same dimension as the input to the corresponding layer in the backbone, so that it could be merged with the input feature to feed the generator progressively.

We perform feature merging in a mask-guided way that it keeps original generator features in the foreground hair region, but replace the background with the output of the encoder. The input feature map to the $i$th last layer in the backbone $F_i$ is blended between the output of the $i+1$th last layer in the backbone $F^g_i$, and the output of the $i$th layer in the condition module $F^b_i$ as:
\begin{equation}
    F_i = F^g_i * M + F^b_i * (1 - M).
\label{eq:background_}
\end{equation}

The input to the background encoder $I_{back}$ is the original background region. We mask out the foreground region by blending the original input $I_{tag}$ with random noise pattern $N$:
\begin{equation}
    I_{back} = N * M' + \mathcal{I}(I * M_{in}, M_{in} - M_{in} \cap M) * (1 - M').
\label{eq:background_}
\end{equation}
Here $M_{in}$ represents the hair mask of the input image $I$. $M'$ indicates the dilated version of the target mask $M$, in order to get rid of remaining hair information in the background due to inaccurate segmentation or stray hair strands. This also implicitly asks the network to refine the thin area $M' - M$ to achieve better boundary blending quality. $\mathcal{I}(I, M)$ is the background image inpainter~\cite{liu2018image} that fills the hole mask $M$ with image content from $I$, in case the edit mask $M$ is smaller than the original hair mask $M_{in}$.

\section{Backbone Generator}
\label{sec:generator}

After introducing all our condition modules, we now present the backbone generation network that integrates all condition inputs to produce the final results, and its training strategy.

\subsection{Architecture}
\label{sec:generator_1}

Given conditional information obtained through the aforementioned condition modules, the generation network, serves as the backbone data flow of the system, aims at producing desired portrait image output that meets all the conditions. According to their particular perceptual characteristics and scales, the outputs of three condition modules are treated in various ways.

As shown in Fig.~\ref{fig:pipelines} (d), our generator is a sequential concatenation of six up-sampling SPADE residual blocks (ResBlk)~\cite{park2019semantic} and a convolutional layer that output the final generation result. As a brief recap of three types of condition inputs:
\begin{itemize}
    \item \textbf{The appearance module} inputs a feature map $A$ to the very first SPADE ResBlk, replacing randomly sampled latent code in traditional GANs;
    \item \textbf{The shape \& structure module} feeds every SPADE ResBlk with stacked hair mask $M$ and orientation map $O$ through the conditional input to modulate the layer activations;
    \item \textbf{The background module} progressively blends background features into the outputs of the last SPADE ResBlks.
\end{itemize}

\begin{figure}[t]
  \centering
  \includegraphics[width=0.75\columnwidth]{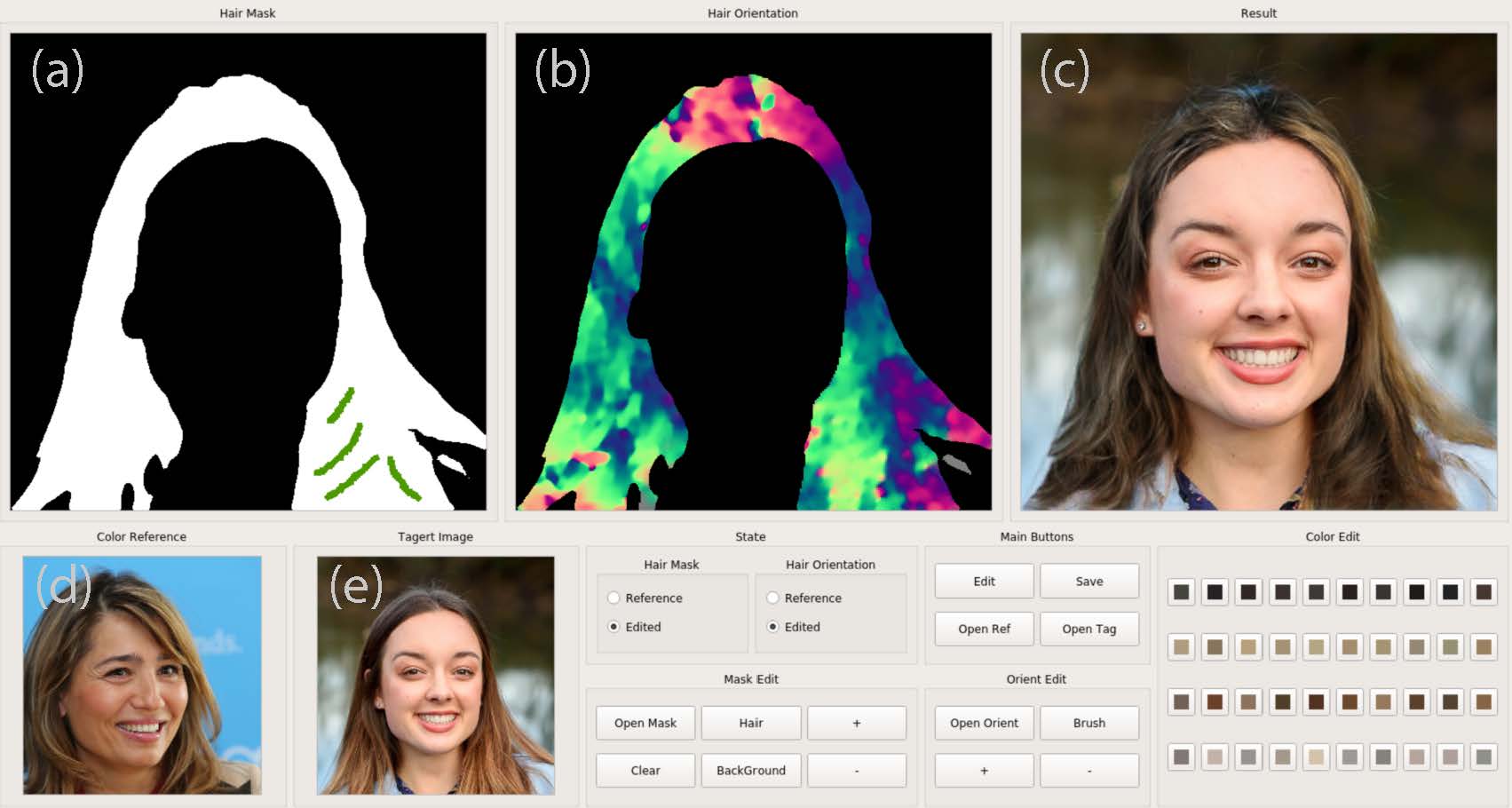}
  \caption{\textbf{Our interactive hair editing system.} (e) shows the original input portrait; (a) is the current shape mask and painting interaction area; (b) is the visualization of current structure orientation map; (d) is the reference appearance photo selected by the user; and (c) is the generated result. Original images courtesy of Deutsche Welle and Loren Kerns.}
  \label{fig:demo}
\end{figure}

\subsection{Training Strategy}
\label{sec:generator_2}

Considering that we do not have the ground truth images when the conditional inputs come from different sources, training the proposed MichiGAN is essentially an unsupervised learning problem. However, most existing unsupervised learning algorithms~\cite{zhu2017unpaired,huang2018multimodal,liu2019few} are usually performing worse than supervised learning regarding the visual quality of the generated results. And more importantly, these unsupervised learning methods are targeted at image-wise translation between different domains, making them less feasible for our specific editing tasks with pixel-wise constraints. Motivated by \cite{he2018deep} which faces a similar problem on image colorization, we propose to train our MichiGAN in a pseudo-supervised way. Specifically, we feed the four conditional inputs extracted from the same source image into MichiGAN and enforce it to recover the original image with explicit supervisions during training, and generalize it to arbitrary combinations of conditions from various sources in inference. Thanks to our hair attribute disentanglement and the powerful learning capacity, MichiGAN gains the generalization ability by leveraging information from all attributes during the supervised training, since ignoring any of them will incur larger losses.

To preprocess the training data, for each training portrait image $I$, we generate its hair mask $M$ with a semantic segmentation network and its orientation map $O$ with the dense orientation estimation method presented in Sec.~\ref{sec:modules_1}. Eventually, we have training pairs with the ground-truth portrait image $I$, and the network output $I' = \mathcal{G}(M, O, I, I)$ that is reconstructed from the conditions.

We adopt the following loss terms to train the network:
\begin{itemize}
    \item \textbf{Chromatic loss.} The generated result should be as close as possible to the ground-truth, but strong pixel reconstruction supervision could also harm the generalization ability of the network. Therefore, instead of measuring full-color reconstruction error, we convert the images into the CIELAB color space and only measure the chromatic distance in \textit{a} and \textit{b} channels, to relax the supervision but still penalize color bias:
\begin{equation}
    L_c = \left \| {\rm Lab}(I')_{\rm ab} - {\rm Lab}(I)_{\rm ab} \right \|_1
\label{eq:background_}
\end{equation}
    \item \textbf{Structural loss.} We propose an additional structure loss to enforce the structural supervision, leveraging our differentiable orientation estimation layer. As described in Sec.~\ref{sec:modules_1}, the orientation estimation layer outputs both orientation $O'$ and confidence $w'$, and the raw orientation output may not be accurate at every pixel. Therefore we formulate the loss as the weighted sum of orientation differences:
\begin{equation}
    L_s = \left \| (O' - O) * w' * M \right \|_1;
\label{eq:orient_loss}
\end{equation}
    \item \textbf{Perceptual loss.} We also measure high-level feature loss with the pre-trained VGG19 model~\cite{simonyan2015very}, which is widely used in previous image translation problems\cite{chen2017stylebank,chen2017coherent,liao2017visual}:
\begin{equation}
    L_p = \left \| {\rm VGG19}(I') - {\rm VGG19}(I) \right \|_1;
\label{eq:orient_loss}
\end{equation}
    \item \textbf{Adversarial loss.} A two-scale PatchGAN~\cite{isola2017image,wang2018high} discriminator $\mathcal{D}$ is used to match distributions between generated results and real portraits to enforce natural visual quality and boost local details;
\commentout{
\begin{equation}
    L_a = \log \mathcal{D}(I) + \log (1 - \mathcal{D}(I'));
\label{eq:orient_loss}
\end{equation}
}
    \item \textbf{Feature matching loss.} To achieve more robust training of GAN, we also adopt the discriminator feature matching loss~\cite{wang2018high}.
\end{itemize}

In summary, the overall training objective can be formulated as:
\begin{equation}
\min_{\mathcal{G}} \max_{\mathcal{D}} (\lambda_c L_c + \lambda_s L_s + \lambda_p L_p + \lambda_a L_a + \lambda_f L_f).
\end{equation}

\begin{table*}[h]
    \caption{\textbf{Frecet inception distance (FID) for different generators and settings.} We use 5,000 images randomly selected from the testing set.}
    \centering
    \begin{tabular}{c|cccc|cccc}
        \hline
        Method & pix2pixHD & SPADE & MaskGAN & MichiGAN & Baseline-NCGA & Baseline-NoS & Baseline-FB & MichiGAN \\
        \hline
         Resolution & \multicolumn{4}{c|}{$512\times512$} & \multicolumn{4}{c}{$256\times 256$} \\
        \hline 
         FID & 21.076 & 21.627 & 49.596 & 10.697 & 31.637 & 31.530 & 30.842 & 21.024 \\
        \hline
    \end{tabular}

    \label{tab:FID}
\end{table*}

\begin{figure}[t]
  \centering
  \includegraphics[width=0.79\columnwidth]{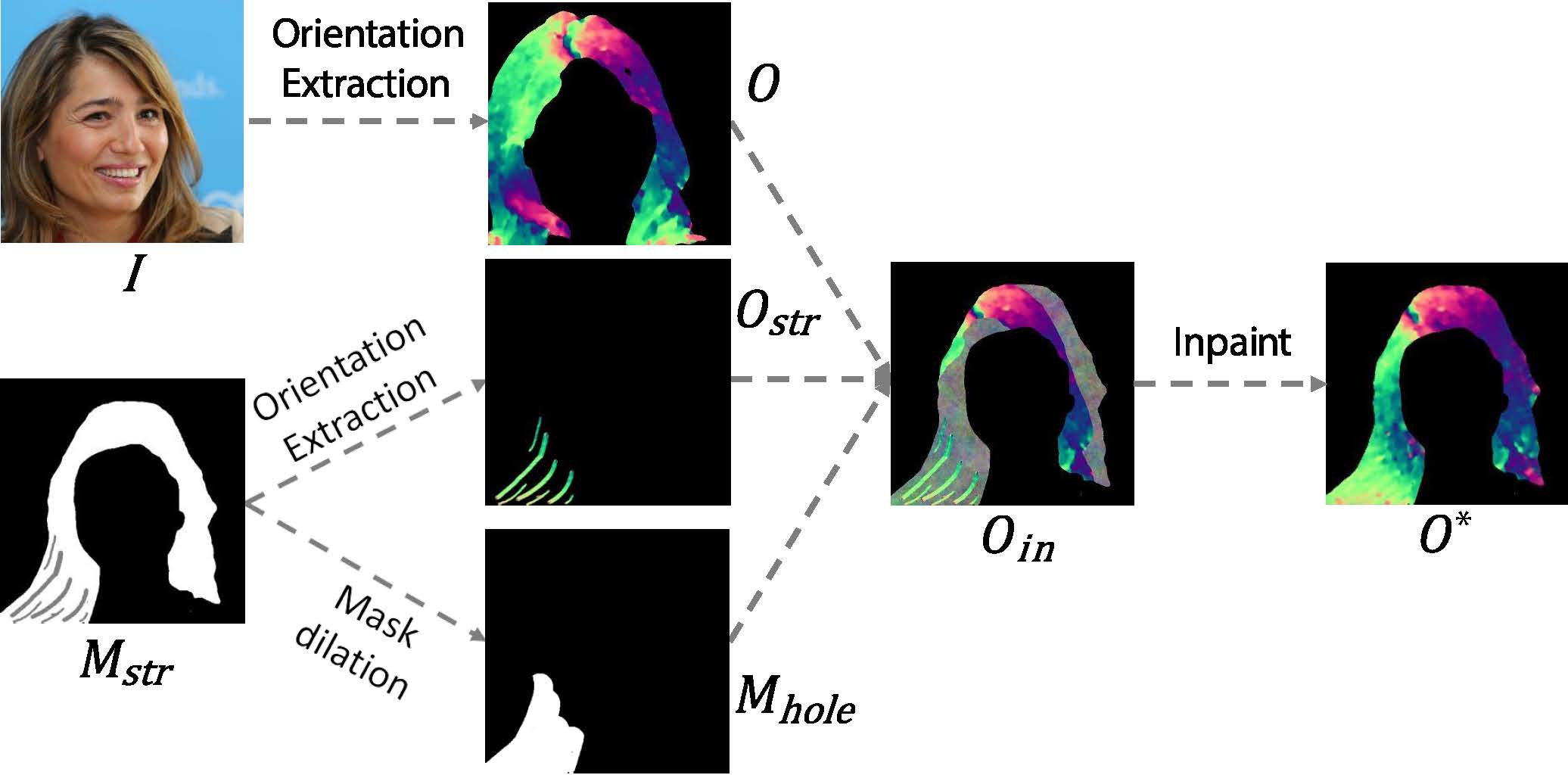}
  \caption{\textbf{Guided orientation inpainting.} Given the current orientation map $O$ extracted from the input image $I$ and user strokes $M_{str}$, we get local hole mask $M_{hole}$ by dilating $M_{str}$ and calculate the guiding orientation $O_{str}$ from $M_{str}$. By merging $O$, $O_{str}$ and noise pattern in $M_{hole} - M_{str}$ together, we obtain the input $O_{in}$ to the inpainting network, which eventually outputs the new orientation map $O^*$ to update the generation result. Original images courtesy of Deutsche Welle.}
  \label{fig:inpaint_orientation}
\end{figure}

\begin{figure*}[t]
  \centering
  \setlength{\tabcolsep}{0\linewidth}
\begin{tabular}{C{0.1\linewidth}C{0.1\linewidth}C{0.2\linewidth}C{0.2\linewidth}C{0.3\linewidth}}
Target&pix2pixHD&SPADE&MaskGAN&MichiGAN\\
\end{tabular}
  \includegraphics[width=0.9\linewidth]{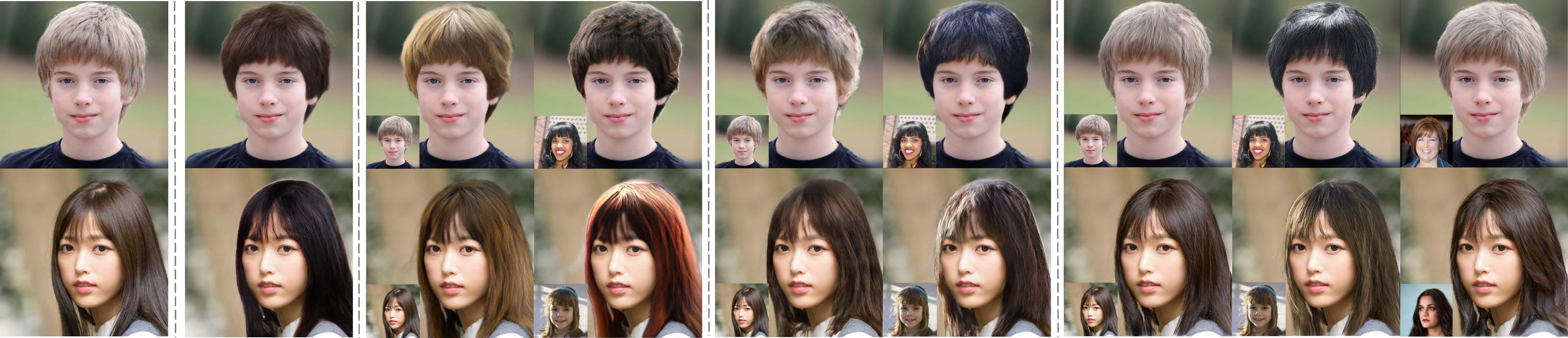}
  \caption{\textbf{Visual comparison with pix2pixHD, SPADE, and MaskGAN.} The hair mask of the original image is used as the input semantic mask for all methods. For SPADE and MaskGAN, a reference image is used to embed the appearance of the hair, which is shown in the left-bottom corner of the result image. For the proposed MichiGAN, we use the same appearance reference image as in SPADE/MaskGAN and take the original image as the orientation reference. In the last column, we also show the example of changing the structure of the hair using MichiGAN with another orientation reference image, which cannot be achieved by previous methods. Compared to these baselines, MichiGAN can produce reliable conditional results with higher visual quality for different condition inputs. Original images courtesy of Tim Wilson, Wilson Pumpernickel, Jeff Latimer, LeafBlue, Bart, and Sophie Marie.}
  \label{fig:sota}
\end{figure*}

\begin{figure*}[t]
  \centering
  \setlength{\tabcolsep}{0\linewidth}
\begin{tabular}{C{0.09\linewidth}C{0.18\linewidth}C{0.18\linewidth}C{0.09\linewidth}C{0.18\linewidth}C{0.18\linewidth}}
Sketch&SC-FEGAN&MichiGAN&Sketch&SC-FEGAN&MichiGAN\\
\end{tabular}
  \includegraphics[width=0.9\linewidth]{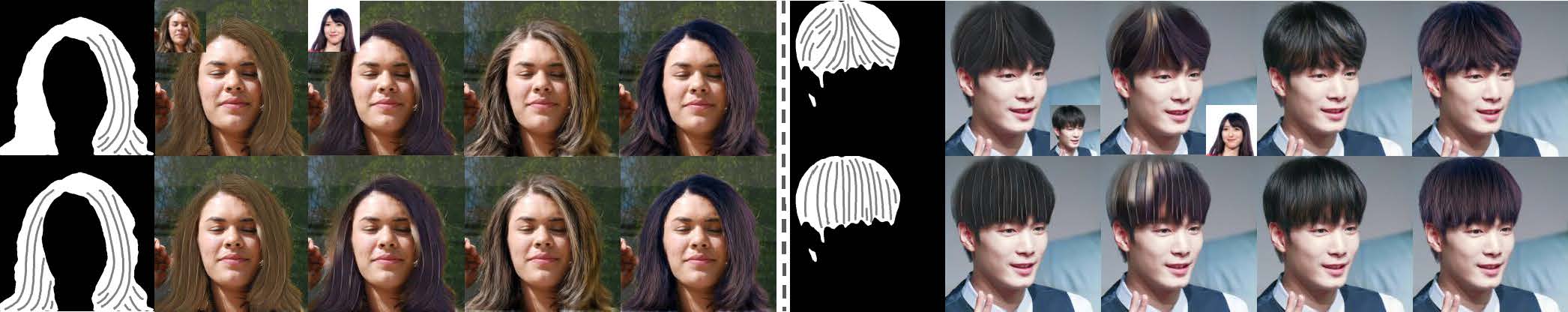}
  \caption{\textbf{Visual comparison with SC-FEGAN.} We use two different user sketches to generate the hair for each target. Two results are generated for each sketch with different appearance references shown in the insets of the SC-FEGAN results, including the image itself (left) and another style reference (right). Original images courtesy of Kevin Burnett, LeafBlue, and butterfly\_60809.}
  \label{fig:vs_scfegan}
\end{figure*}

\begin{figure*}[t]
  \centering
\setlength{\tabcolsep}{0\linewidth}
\begin{tabular}{C{0.1\linewidth}C{0.1\linewidth}C{0.1\linewidth}C{0.1\linewidth}C{0.1\linewidth}C{0.1\linewidth}C{0.1\linewidth}C{0.1\linewidth}C{0.1\linewidth}}
Target&Structure&Result&Target&Appearance&Result&Target&Shape&Result\\
\end{tabular}
  \includegraphics[width=0.9\linewidth]{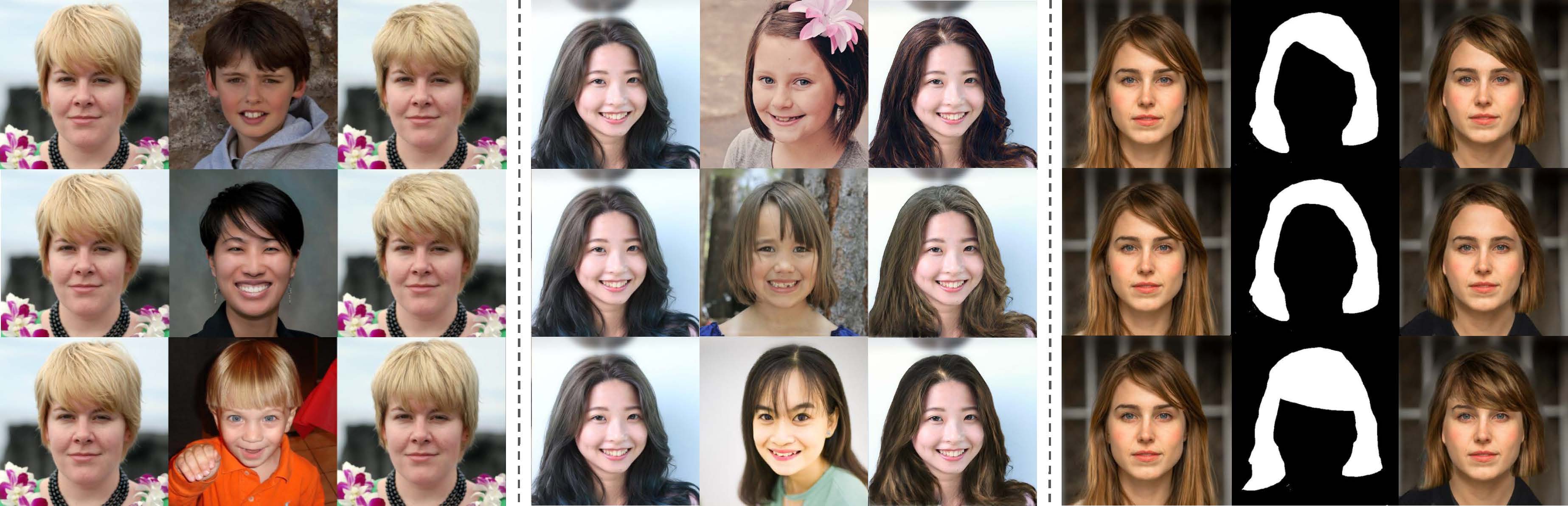}
  \caption{\textbf{Reference-based hair editing results with a single attribute adjusted.} It shows that MichiGAN disentangles these visual attributes well without the interference of each other, thus enabling editing each attribute individually. Original images courtesy of Joshua Ganderson, Owen Lucas, Unity Health Care, Stonewall Ballard, VcStyle, Joanna Layrock, thejerry, NGÁO STUDIO, and Thomas Hawk.}
  \label{fig:result2_3}
\end{figure*}

\section{Interactive Editing System}
\label{sec:system}

Based on the proposed MichiGAN, we also build an interactive portrait hair editing system. Fig.~\ref{fig:demo} shows one screenshot of the user interface of the editing system. Within this system, users can control the shape (top left), structure (top middle), and the appearance (bottom left) separately or jointly, and see the result at the top right.

To facilitate user control over multiple approaches, we further propose two types of interaction modes, namely reference mode and painting mode.

\paragraph{Reference mode}
One direct motivation for a user to manipulate the hair in a portrait, is to try hair attributes from another reference photo onto the target one. So we provide support to such an intuitive interaction approach by allowing the user to offer reference portraits for different attributes individually or jointly, and extracting network-understandable condition inputs to guide the generation.

Specifically, users are allowed to specify up to three references for shape ($I_{shape}$), structure ($I_{str}$), and appearance ($I_{ref}$), respectively. The system then calculates hair mask $M$ from $I_{shape}$, orientation map $O$ from $I_{str}$, and uses $I_{ref}$ to extract the appearance features. These inputs are used to drive the generation of results.

When the target mask and orientation map come from different sources, the shape of the orientation may not fully cover the entire target shape. Therefore, we train an orientation inpainting network similar to the background one (Sec.~\ref{sec:modules_3}) to complete the missing holes in the orientation map.

\paragraph{Painting mode}
It is also natural to let a user perform local and detailed editing to the image with a much higher degree of flexibility than the reference mode. To achieve so, we provide a stroke-based interaction interface to enable local modification to both shape and structure, and a color palette tool to control the hair appearance. Their underlying methods are introduced below:
\begin{itemize}
    \item \textbf{Shape painter.} Users can change the binary hair mask with shape painter to augment or remove certain mask regions at will, similar to the interface of~\cite{park2019semantic}. The smooth boundaries of these edited regions may not match true hair boundary quality, but our fuzzy mask augmentation (Sec.~\ref{sec:modules_1}) could still nicely capture these details.
    \item \textbf{Structure painter.} Structure interaction is a way more challenging than the shape, since orientation is dense and not intuitively understandable. Regarding that, we propose a guided structure editing method to enable user-friendly manipulation. Given the current orientation map $O$ and a set of user guidance strokes $M_{str}$, within a certain local region $M_{hole}$ around the strokes, we synthesize new orientation information, which should be compatible with both the stroke guidance and the outside regions. The hole mask $M_{hole}$ is generated by dilating the thin mask of user strokes with a certain radius. And the guidance orientation $O_{str}$ with the stroke is calculated with the known painting orders of line segments that form the stroke paths. We compose all the information into a partial orientation map $O_{in}$ as:
\begin{equation}
    O_{in} = O * (1 - M_{hole}) + O_{str} * M_{str} + N * (M_{hole} - M_{str}),
\end{equation}
    where $N$ represents a random noise pattern to hide the original orientation. We use an orientation inpainting network to synthesize new orientation inside the hole region $M_{hole}$, which is trained on paired data of ground-truth orientation maps and synthesized random user strokes traced on them.
    \item \textbf{Appearance picker.} Finding a specific good appearance reference could be a headache for users. In painting mode, we design a palette-like appearance picker tool to retrieve, cluster, and navigate references by annotating a single RGB hair color. This is enabled by embedding all our reference portraits (selected from our test dataset) into the color space by calculating the average hair color. Given an arbitrary target color, we do KNN search around it to find $K$ closest references for the user to test. And once the user picks a reference, the system will instantly update the candidate list to contain references clustered around the selected one.
\end{itemize}

\section{Experiments}

\subsection{Implementation Details}
\paragraph{Datasets.}
To generate data pairs for supervised training, we use the large-scale portrait dataset of Flickr-Faces-HQ (FFHQ)~\cite{karras2019style}. The whole dataset is divided into two parts: $56000$ images for training and $14000$ images for testing. The resolution of images is resized to $512\times512$.

\paragraph{Network Structures.}
We illustrate the structures of MichiGAN in Fig.~\ref{fig:pipelines}. Specifically, the appearance module uses five consecutive downsampling partial convolutions followed by an instance average pooling to get the appearance vector. All these partial convolutions are masked by the appearance reference, with the same kernel size of $3\times3$ and increasing feature channels of $64/128/256/512/1024$. Each partial convolution is followed by an instance normalization and a leaky ReLu activation. A mask transform block is then used to spatially duplicate the appearance feature vector to an appearance map (size $8\times8$), guided by the target hair mask. The shape and structure modules follow the same modulation networks in SPADE to denormalize each SPADE ResBlk in the backbone with target hair mask and orientation (after inpainting). The background module generates the background input via inpainting and noise-filling, and produces $4$ scales of background features using a network that consists of four $3\times3$ convolution layers with feature channels of $64/128/256/512$. These features are blended into the backbone guided with the background mask. Finally, the backbone generator accepts the appearance map as input, and goes through six upsampling SPADE ResBlks and one final convolution to generate the result. The feature channels of these layers are $1024/1024/1024/512/256/128/64$.
The inpainting networks for both orientation and background follow the structure proposed by~\cite{liu2018image}.

\paragraph{Training Parameters.}
We train all condition modules and the backbone generator jointly. The Adam optimizer \cite{kingma2014adam} is used with a batch size of $8$ and the total epoch number of $50$. The learning rates for the generator and discriminator are set to $0.0001$ and $0.0004$, respectively. The structural loss weight $\lambda_s$ is $10$, while all other loss weights, $\lambda_c$, $\lambda_p$, $\lambda_a$, and $\lambda_f$ are simply set to $1$ (Sec.~\ref{sec:generator_2}). To obtain the dilated target mask $M'$ (Sec.~\ref{sec:modules_3}), $M$ is randomly expanded by $[0.03, 0.07]$ of the image width.

\subsection{Comparison with SOTA}

\paragraph{Semantic Image Generation.}
We compare MichiGAN with two semantic image synthesis baselines: pix2pixHD~\cite{wang2018high} and SPADE~\cite{park2019semantic}. Pix2pixHD only supports generation based on semantic masks. And SPADE allows user control over both the semantic mask and the reference style image, which is the current state-of-the-art GAN-based conditional image synthesis framework. We train both pix2pixHD and SPADE models using the implementations provided by their authors, and exactly the same data as we used to train MichiGAN. For a fair comparison, the hair mask of the original image is converted to the input semantic mask while the reference image is set to be the original input. Besides, since pix2pixHD and SPADE can not maintain the background unchanged, we blend their generated hair with the original background with the soft hair mask. In this reconstruction setting, the original image can be regarded as the ground truth of the desired result, since all the input conditions are derived from it.

Quantitatively, MichiGAN outperforms the competing methods by a large margin, as shown in Tab.~\ref{tab:FID}. Our method achieves an FID score of $10.7$, which is less than half of pix2pixHD and SPADE. It indicates the edited portrait images by our method have much better realism than these two baseline methods.

Qualitatively, overall, all the methods can produce plausible hair results while our results have the best visual quality. A set of visual comparison results are shown in Fig.~\ref{fig:sota}. Taking a closer look at the details of synthesized hair, we can find that the results generated by pix2pixHD and SPADE models are coarse-grained and have some apparent artifacts, while our results are more fine-grained with the rich and delicate texture details.

In terms of controllability, MichiGAN performs much better than pix2pixHD and SPADE. 
Pix2pixHD can only generate hair that looks adaptive to the input hair mask but with uncontrollable color and structure. SPADE can also synthesize hair adaptive to the semantic mask.
Compared to pix2pixHD, it has additional ability to transfer the style of the reference image to the synthesized hair.
However, the transferred results look apparently different from the desired ones, even using the original image as the reference, which is unsatisfactory. Besides, SPADE can not control the structure of the synthesized hair either.
Compared to pix2pixHD and SPADE, MichiGAN can not only produce hair adaptive to the semantic mask, but also precisely control the different attributes of the hair, including appearance and structure, so that we can produce desired hair that looks most similar to the references.
\commentout{When both the appearance reference and the structure reference are set as the original image, our method can reconstruct the original image almost perfectly.}

In addition, the hair produced by pix2pixHD and SPADE is not compatible with the background of the original image. Despite that the results are softly blended with the backgrounds using high-quality hair masks, we can still easily see the stitching artifacts around the hair boundaries, especially for some complicated backgrounds, which introduces difficulties when applying pix2pixHD and SPADE to interactive editing. In contrast, MichiGAN does not suffer from this problem, since we take the background of the original image as one of the input conditions to directly generate the entire image with both the desired hair and the fixed background.

\paragraph{Conditional Face Generation.}
We also compare MichiGAN with two recent conditional face generation methods: MaskGAN~\cite{lee2019maskgan} and SC-FEGAN~\cite{jo2019sc}.

MaskGAN uses fine-grain facial semantic masks as the representation to manipulate face images and is able to transfer styles between different subjects, which is similar to SPADE~\cite{park2019semantic}. Therefore we conduct the comparison with MaskGAN in the same mask-conditioned generation experiments, as shown in Fig.~\ref{fig:sota} and Tab.~\ref{tab:FID}. As can be seen from both qualitative and quantitative results, although MaskGAN can synthesize visually plausible hair, our method still achieves better visual quality while being able to preserve or control the structural details.

SC-FEGAN adopts free-form sketches for portrait generation in the target hole regions to control both structure and appearance, which is comparable to our stroke-based hair generation. For a fair comparison, we adopt the hair mask as the target hole region, and use a same set of user strokes painted with our interactive system. The strokes are converted to the sketch input to SC-FEGAN, together with the corresponding hair color samples. A set of visual results are shown in Fig.~\ref{fig:vs_scfegan}, which contains comparisons on two target images with different sketches and appearance references. As demonstrated by, although SC-FEGAN allows control over hair structure and color to some extent, the generated results are less realistic comparing with ours. This may be partially caused by the gap between synthetic sketches generated by an edge detector and real user strokes. Our method avoids such a gap by using the dense orientation map as a unified representation for the hair structure.

\subsection{Qualitative Results}
Our interactive hair editing system based on MichiGAN supports flexible manipulation of hair attributes, including shape, structure, and appearance, individually or jointly. These attributes can be conveniently specified in two ways, either by giving a reference image or interactively modifying original ones with painters, which corresponds to two modes of the system: reference-based editing and painting-based editing.

\begin{figure}[t]
  \centering
\setlength{\tabcolsep}{0\linewidth}
\begin{tabular}{C{0.2\linewidth}C{0.2\linewidth}C{0.2\linewidth}C{0.2\linewidth}C{0.2\linewidth}}
Target&Appearance&Structure&Shape&Result\\
\end{tabular}
  \includegraphics[width=\linewidth]{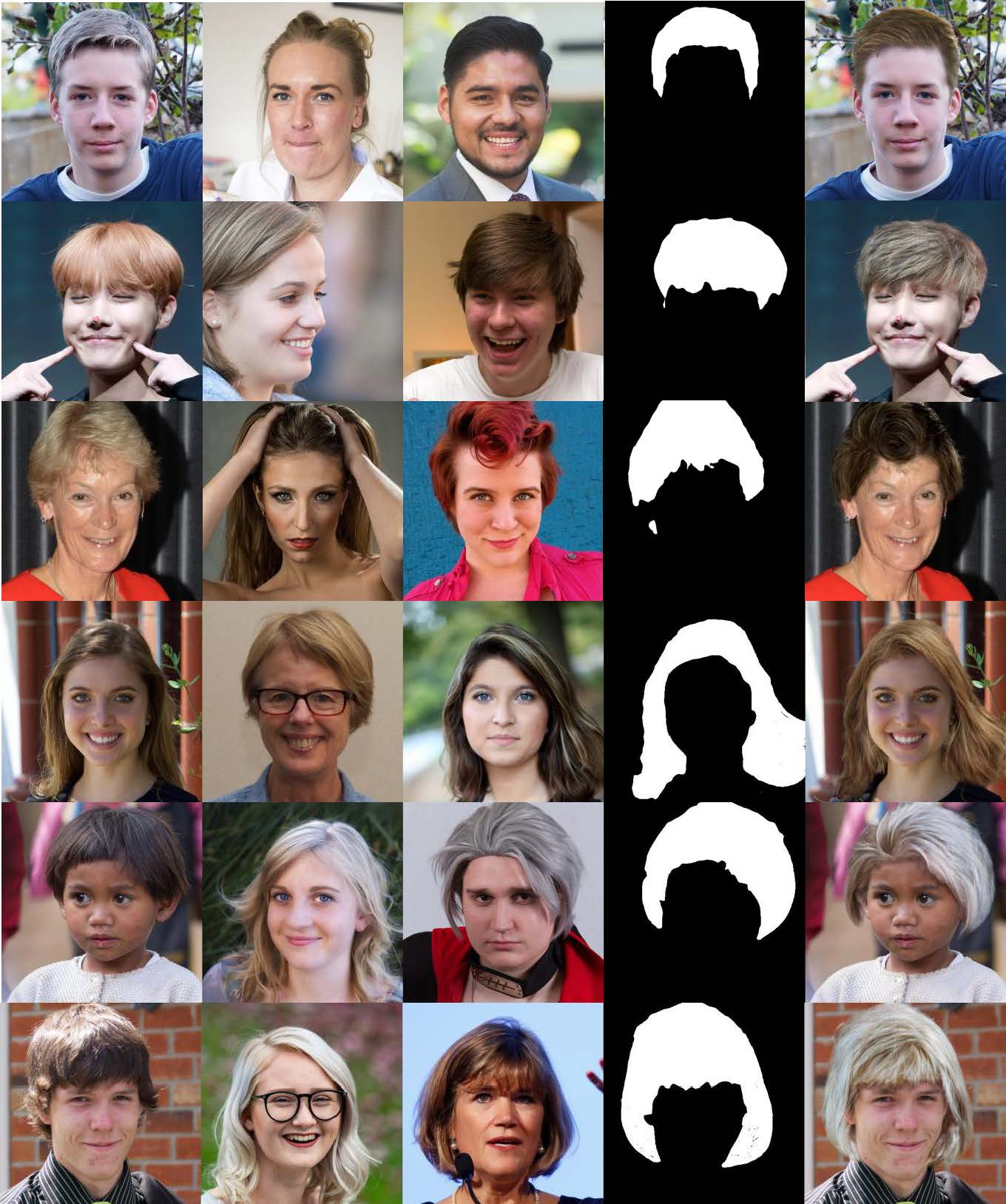}
  \caption{\textbf{Reference-based hair editing results with multiple attributes adjusted jointly.} MichiGAN enables editing multiple attributes with each attribute from one different reference image. Original images courtesy of Dean Chatterson, iwishmynamewasmarsha, Elio Yañez, Beautiful Star, John Benson, Alex Luyckx, Neville Wootton, Flavio\~, Megan Rosenbloom, stephen davis, Eric E Castro, Pawel Loj, RAFFI YOUREDJIAN, Devon Christopher, Wm Andrew Murphy, Keven Menard Photography, University of the Fraser Valley, and sydsmomma24.}
  \label{fig:result1}
\end{figure}

\paragraph{Reference-based results.}  
We first show our reference-based hair editing results on in-the-wild portrait images. In this mode, the user first loads an original image into our system and then can simply adjust an attribute by providing another reference image. For example, once a  reference image is given for the structure attribute, the orientation map will be extracted automatically by our system, and an image with a new hair structure will be generated based on the orientation condition. It is the same for appearance and shape.  Figure~\ref{fig:result2_3} show the results with each attribute adjusted individually, while Figure~\ref{fig:result1} shows the results with multiple attributes adjusted jointly. It can be observed that MichiGAN disentangle these visual hair attributes well, without the interference of each other. And all these attributes are naturally embedded in generating a new image.

\begin{figure*}[t]
  \centering
  \includegraphics[width=0.9\linewidth]{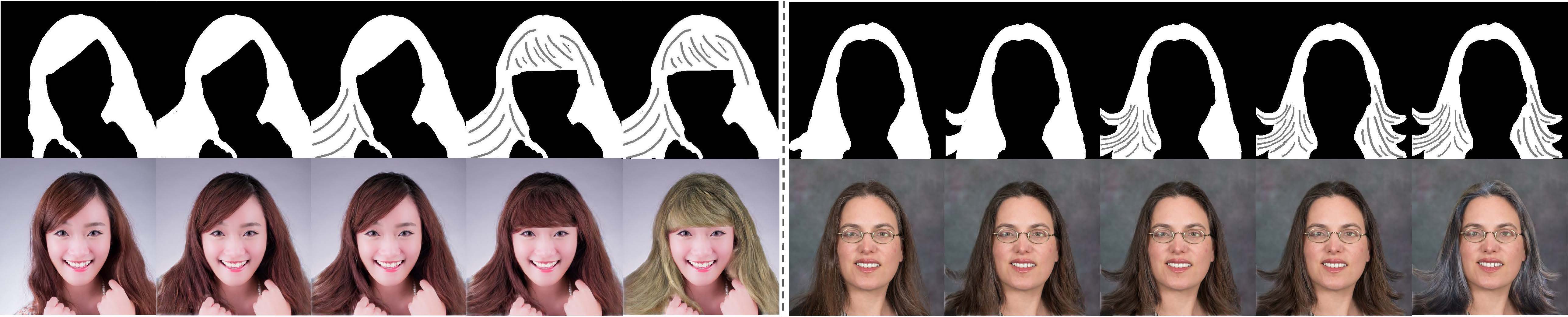}
  \caption{\textbf{Hair editing results with our interactive system.} Users can use a brush tool to paint the mask shape, modify the structure with orientation strokes, or change the appearance by picking an alternative color. Original images courtesy of IOTA[Ext] and UC Davis College of Engineering.}
  \label{fig:interactive}
\end{figure*}

\paragraph{Painting-based results.} 
In this part, we show painting-based hair editing results, where the attributes are modified by the user with an interactive painting interface (Fig.~\ref{fig:demo}). Compared to the reference-based mode, this painting-based mode is more natural and flexible for the user to perform some local and detailed edits. Two examples are given in Fig.~\ref{fig:interactive}. The user first uses a brush tool to modify the mask image, thus the shape of the hair is changed accordingly. Then several strokes are added to indicate the orientation of hair, and a new structure of hair is generated to follow the orientations. Finally, the user changes the appearance of the hair by specifying some fancy colors.  Please refer to the accompanying video for a demonstration. Since our interface is user-friendly and the generated result is intermediately fed back to the user, editing hair with our system is quite convenient. The user study shows that 15-30 seconds on average is spent for an amateur to edit one example.

\begin{figure}[t]
  \centering
\setlength{\tabcolsep}{0\linewidth}
\begin{tabular}{C{0.25\linewidth}C{0.25\linewidth}C{0.25\linewidth}}
Hair Reference&Target&Result\\
\end{tabular}
  \includegraphics[width=0.75\linewidth]{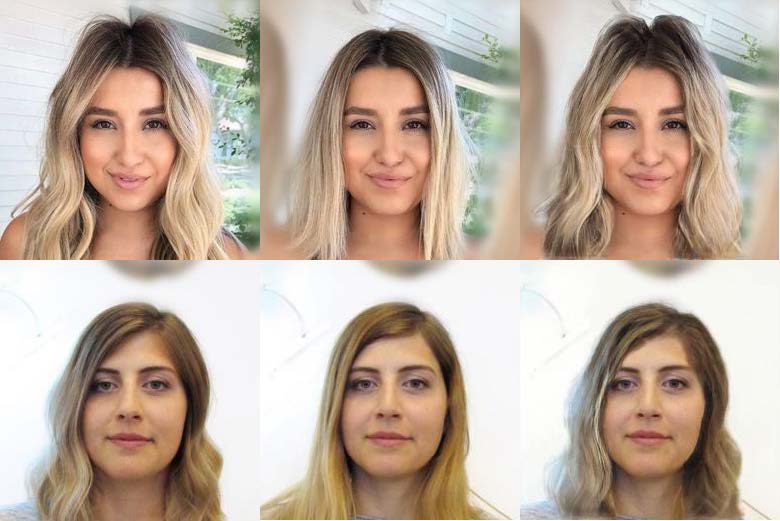}
  \caption{\textbf{Hairstyle Transfer Validation.} We collect two photos of a same person with different hairstyles (left and middle). The result (right) is generated by using one image (left) as the reference that provides hair conditions, and the other one (middle) as the target that provides the background. Original images courtesy of Bob HARRIS.}
  \label{fig:self}
\end{figure}

\paragraph{Hairstyle Transfer Validation.}
We also validate the realism of our results by transferring the hairstyle from one image to another on a same subject. As shown in Fig.~\ref{fig:self}, we collect two images with different hairstyles of a same person, and use one image as the source of hair conditions, and the other one as the target to generate a hairstyle transfer result. Our method achieves realistic results with both appearance and structure similar to the ground truth photo.

\begin{figure}[th!]
  \centering
\setlength{\tabcolsep}{0\linewidth}
\begin{tabular}{C{0.2\linewidth}C{0.2\linewidth}C{0.2\linewidth}C{0.2\linewidth}}
Target&Appearance&NCGA&MichiGAN\\
\end{tabular}
  \includegraphics[width=0.8\linewidth]{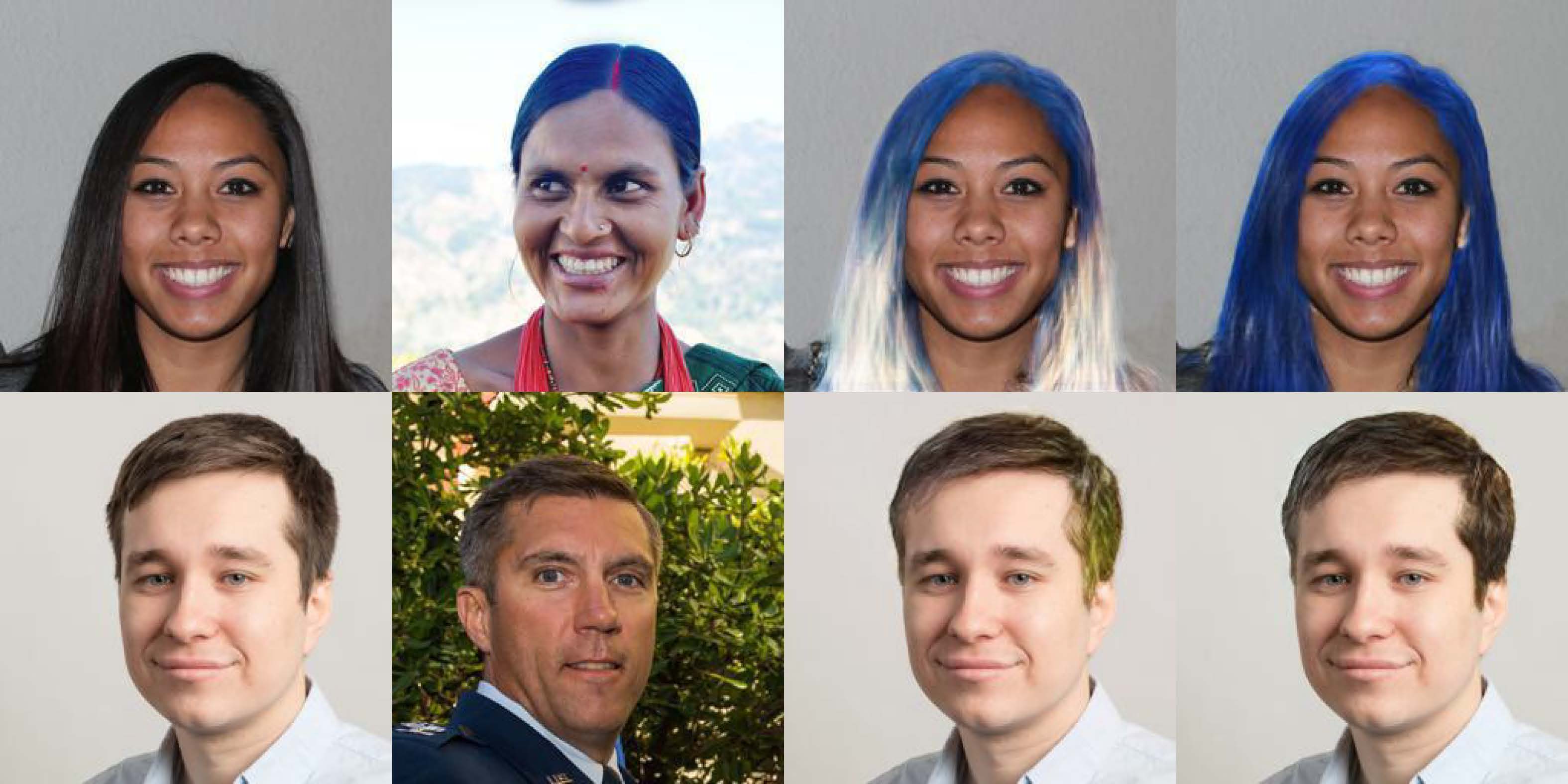}
  \caption{\textbf{Comparisons with the baseline model using normal convolution and global average pooling.} Baseline-NCGA tend to absorb the background color into the appearance features while our method does not. Original images courtesy of ErrysF, Possible, NTNU Faculty of Natural Sciences, and Anthony Quintano.}
  \label{fig:model1}
\end{figure}

\subsection{User Study}
To further evaluate the realism to real human perception, a simple user study is conducted on Amazon Mechanical Turk. We prepare two image sets, each has $200$ images. One contains real portrait photos, and the other contains results generated by our method with condition inputs from randomly selected test images. During the study, we mix both sets and randomly show one image at a time to a user, and ask the user to subjectively judge whether this image is real or fake. Each image has been evaluated by $3$ users. Each user spends $3$ seconds on each image on average. The final fooling rates of our results and real images are $64.67\%$ and $81.10\%$, respectively.

\subsection{Ablation Study}
In order to achieve plausible results, we dedicatedly design the proposed framework with several key ingredients, ranging from the network architecture design to the newly introduced objective functions. But considering the space limitation and training cost, only three key ablation analysis experiments are conducted, and the image resolution is set to $256\times 256$.

\paragraph{Partial Convolution and Instance-wise Average Pooling.} 
When adopting the reference image as the appearance guidance, we only want the appearance condition module to absorb the appearance information from the hair region but not the background region. 
But this is non-trivial because most existing operators in modern neural networks do not naturally support it. For example, the normal convolutional layer processes the pixels within a regular grid for all the possible sliding windows. Considering the hair region is often of an irregular shape, directly applying these normal operators in the appearance module will inevitably introduce the background information into the encoded appearance features. Based on this observation, we replace all the normal convolutional layers and average pooling layers with partial convolutional layers \cite{liu2018image} and instance-wise average pooling \cite{wang2018high} respectively.

To show its necessity, we conduct an ablation experiment that adopts normal convolutional and global average pooling layers, denoted as ``Baseline-NCGA". Obviously, from Fig.~\ref{fig:model1}, we can find that the generated hair of the baseline model has significantly introduced the background color and becomes very unrealistic. Specifically, in these two cases, the target hair is polluted by the white color from the background sky and the green color from the background tree, respectively. By contrast, our method only adopts the reference hair color in the final generated results. 
As shown in Tab.~\ref{tab:FID}, our method also achieves a better FID score than Baseline-NCGA.

\begin{figure}[h!]
  \centering
\setlength{\tabcolsep}{0\linewidth}
\begin{tabular}{C{0.2\linewidth}C{0.2\linewidth}C{0.2\linewidth}C{0.2\linewidth}C{0.2\linewidth}}
Target&Appearance&Structure&NoS&MichiGAN\\
\end{tabular}
  \includegraphics[width=\linewidth]{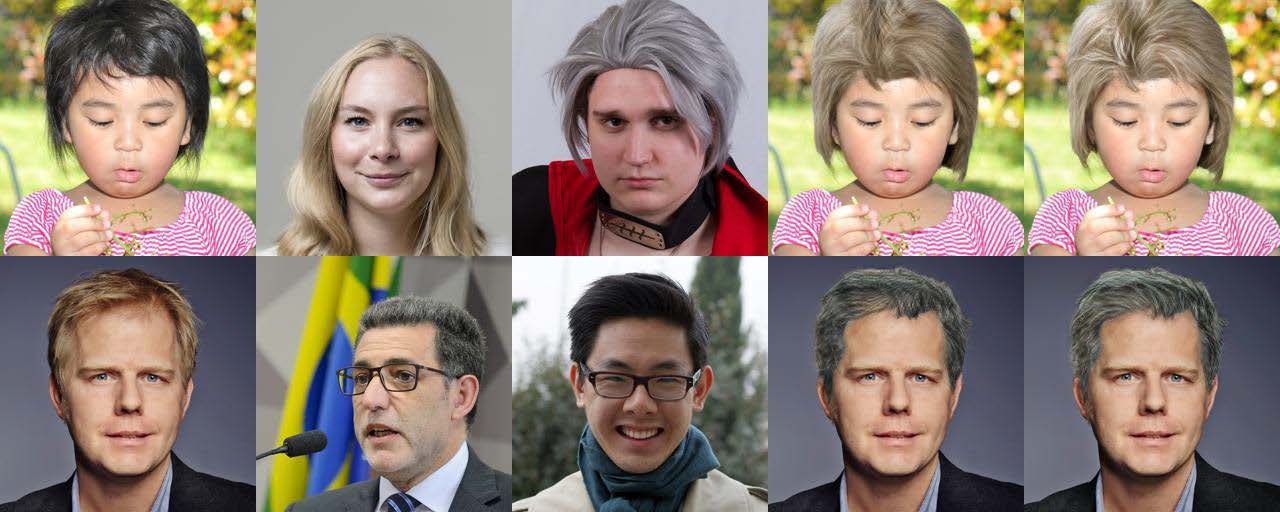}
  \caption{\textbf{Comparisons with the baseline model without structural loss.} It can be seen that Baseline-NoS cannot adopt the reference hair orientation in the final generated target hair. Original images courtesy of musetheory, rødt.no, Wm Andrew Murphy, Thomas Benkö, Senado Federal, and PriceMinister.}
  \label{fig:model2}
\end{figure}

\paragraph{Necessity of Structural Loss.} To enable structural hair control, besides the dense orientation map as the condition input, we also leverage one differentiable orientation layer in the network and add an extra structural loss as training supervision. In this ablation experiment, we are curious about whether the structural loss is necessary. To verify it, we still feed the orientation map into the network but ignore the structural loss term in the training objective. 
As shown in Tab.~\ref{tab:FID}, our method performs better than this baseline in terms of the FID score. 
In Fig.~\ref{fig:model2}, we provide two typical examples where both the appearance and orientation guidance are given by a reference image. It can be seen that, though the baseline method can adopt the appearance from one reference image, it cannot enable the structure control at all. Therefore the hair orientation in its results is very orderless and makes the overall result not realistic enough. From this experiment, two conclusions can be drawn: 1) Hair structure control is not only a feature but also a key component for realistic hair generation; 2) Without explicit structural supervision, it is difficult for the generative network to generate sensible hair orientation, let alone keeping the reference orientation. 

\begin{figure}[t]
  \centering
\setlength{\tabcolsep}{0\linewidth}
\begin{tabular}{C{0.2\linewidth}C{0.2\linewidth}C{0.2\linewidth}C{0.2\linewidth}C{0.2\linewidth}}
Target&Appearance&Structure&FB&MichiGAN\\
\end{tabular}
  \includegraphics[width=\linewidth]{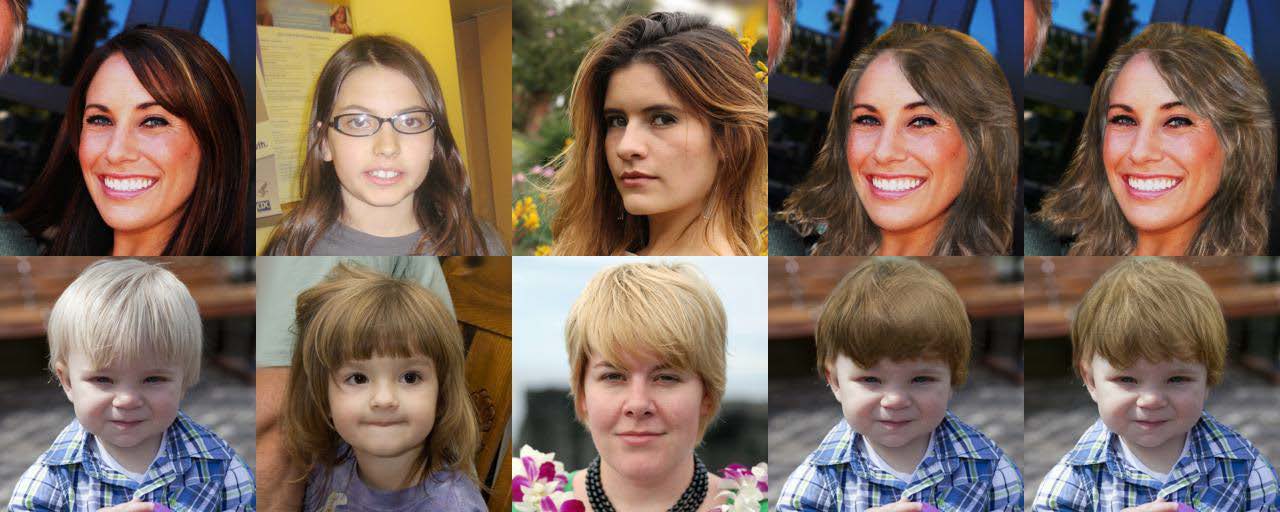}
  \caption{\textbf{Comparisons with the baseline that uniformly blends the foreground and background feature without the mask guidance.} Compared to our method, Baseline-FB cannot preserve the reference appearance and orientation well due to the interference of the background. Original images courtesy of Ralph Zuranski, RTLibrary, Julia Suárez Luna, evan courtney, jstonkatoy, and Joshua Ganderson.}
  \label{fig:model3}
\end{figure}

\paragraph{Mask-Guided Feature Blending.} Though the main goal of our method is foreground hair editing, how to keep the original background unchanged and blend it with edited hair without artifacts is still very crucial. In order to address this problem, we choose to adopt a background condition module and fuse the background feature with the foreground feature in a mask-guided way in this paper. To show the necessity of the mask guidance, we further compare our method with the naive baseline that directly blends the foreground and background features with the mask guidance. In Fig.\ref{fig:model3}, two representative examples are provided.  Obviously, this naive blending way will introduce the background feature into the foreground region, and cannot preserve the reference appearance and orientation well. 
The better FID score shown in Tab.~\ref{tab:FID} further verifies the necessity of our design.

\section{conclusion}
\begin{figure*}[t]
  \centering
\setlength{\tabcolsep}{0\linewidth}
\begin{tabular}{C{0.1\linewidth}C{0.1\linewidth}C{0.1\linewidth}C{0.1\linewidth}C{0.1\linewidth}C{0.1\linewidth}C{0.1\linewidth}C{0.1\linewidth}C{0.1\linewidth}}
Target&Shape&Result&Target&Appearance&Result&Target&Strokes&Result\\
\end{tabular}
  \includegraphics[width=0.9\linewidth]{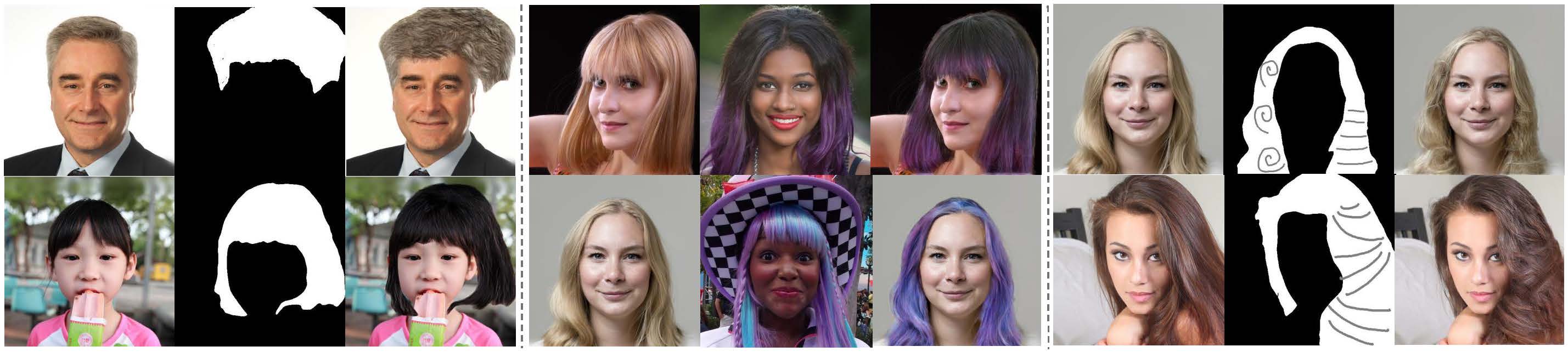}
  \caption{\textbf{Three possible limitations of MichiGAN}: 1) we need the target mask to be aligned reasonably well with the face; 2) we cannot capture the spatially varying appearance well; 3) the painted stroke paths cannot be too extreme. Original images courtesy of International Hydropower Association, rox.chen, begemot\_dn, Jay Weenig, Jennie Park  mydisneyadventures, rødt.no, and barefashion.}
  \label{fig:limitations}
\end{figure*}

To conclude, we present MichiGAN, a conditional hair generation network for portrait manipulation applications. Different from existing conditioned face generation methods, our method provides user control over every major hair visual factor, which is achieved by explicitly disentangling hair into four orthogonal attributes, including shape, structure, appearance, and background. For each of them, we modulate the image generation pipeline with a specially designed condition module that processes user inputs with respect to the unique nature of the corresponding hair attribute. All these condition modules are integrated with the backbone generator to form the final end-to-end network, which allows fully-conditioned hair generation from various user inputs. Based on it, we also build an interactive portrait hair editing system that enables straightforward manipulation of hair by projecting intuitive user inputs to well-defined condition representations.

Despite the superior performance of our MichiGAN, there still exist several limitations to be explored in the future (Fig.~\ref{fig:limitations}):
\begin{itemize}
    \item The target shape mask, either painted or from a reference, should be aligned reasonably well with the face. The mismatched mask can result in unnatural hair shape. It would be of great help if we can automatically warp the hair shape according to different target face poses. We plan to investigate this topic in future research.
    \item The appearance module generally works well on hair images with a globally consistent appearance. However, spatially-varying hair appearance might be smeared out in the results.
    \item Even with our guided orientation inpainting, the guidance stroke should still be naturally compatible with the shape. Extreme stroke paths may cause unexpected structural issues since they can hardly be found in the synthetic training pairs.
    \item We randomly dilate/erode hair shape masks and shrink the background masks during training to enforce a certain ability to synthesize natural hair boundaries from binary masks. However, the results can still be less satisfactory when hair shapes are dramatically changed, since we do not explicitly handle boundary matting. In the future, we are interested in combining MichiGAN with the latest learning-based alpha-matting methods to further increase the quality.
\end{itemize}

\section*{acknowledgements}
We would like to thank the reviewers for their constructive feedback, Mingming He and Jian Ren for helpful discussions, Nvidia Research for making the Flickr-Faces-HQ (FFHQ) dataset, and all Flickr users for sharing their photos under the Creative Commons License. This work was partly supported by Exploration Fund Project of University of Science and Technology of China, YD3480002001 and Hong Kong ECS grant \#21209119, Hong Kong UGC.
 
\bibliographystyle{acmsiggraph}
\nocite{*}
\bibliography{acmsiggraph}
\end{document}